\documentclass[runningheads]{llncs}
\usepackage[T1]{fontenc}
\usepackage{graphicx}
\usepackage{tikz}
\usepackage{lingmacros}
\usepackage{tree-dvips}
\usepackage{pgfplots}
\usepackage{xlop}
\pgfplotsset{width=7cm,compat=1.9}
\usepackage{scalerel}
\begin{document}
\title{Transformer-based Parameter Estimation in Statistics}
%
%
\author{Xiaoxin Yin\inst{1}
             David S. Yin\inst{1}}
\institute{Independent Researchers\\
	San Jose, California, USA\\
	\email{\{xiaoxin,davidsyin\}@gmail.com}}
\maketitle              
\begin{abstract}
Parameter estimation is one of the most important tasks in statistics, and is key to helping people understand the distribution behind a sample of observations. Traditionally parameter estimation is done either by closed-form solutions (e.g., maximum likelihood estimation for Gaussian distribution), or by iterative numerical methods such as Newton-Raphson method when closed-form solution does not exist (e.g., for Beta distribution). 

In this paper we propose a transformer-based approach to parameter estimation. Compared with existing solutions, our approach does not require a closed-form solution or any mathematical derivations. It does not even require knowing the probability density function, which is needed by numerical methods. After the transformer model is trained, only a single inference is needed to estimate the parameters of the underlying distribution based on a sample of observations. In the empirical study we compared our approach with maximum likelihood estimation on commonly used distributions such as normal distribution, exponential distribution and beta distribution. It is shown that our approach achieves similar or better accuracy as measured by mean-square-errors. 

\end{abstract}
\section*{1. Introduction}
Parameter estimation is one of the most important tasks in statistics. Given the type of the underlying distribution, and a sample of observations, a parameter estimation method should produce an estimate of the distribution’s parameters. Taking 1-dimensional normal distribution as an example. Given a sample $x_1$, …, $x_N$ drawn from normal distribution $\mathcal{N}(\mu,\sigma^2)$, we hope to estimate its parameters $\mu$ and $\sigma$. The most commonly used method is MLE (maximum likelihood estimation). It can be easily proven that the MLE of $\mu$ is $\hat{\mu} = \frac{\sum_{i=1}^{N} x_i}{N}$, and that of $\sigma$ is $\hat{\sigma} = \sqrt{\frac{\sum_{i=1}^{N} {(x_i - \hat{\mu}})^2}{N}}$. The proof can be found in~\cite{c:17}. 

Parameter estimation for some other distributions can be much more complex. For example, it is shown that the maximum likelihood estimation for the parameters of a Beta distribution has no closed form~\cite{c:08}, and one needs to resort to numeric optimization methods such as Newton-Raphson method or gradient descent to estimate its parameters.

In recent years transformer-based approaches have been successfully applied to many diverse tasks, including many math tasks such as solving word problems~\cite{c:21}, mathematical reasoning~\cite{c:23}, symbolic regressions~\cite{c:20}, and time-series forecasting~\cite{c:20-2}. In this paper we propose an approach to perform parameter estimation using deep learning. For each type of distribution, we generate a training set, with each training example containing a sample drawn from a distribution with predetermined parameters. Then we convert each sample into a sequence of embeddings, and train a transformer to predict the parameter(s) with the sequence being the input. We aim at producing estimations that are as close as possible to the true parameters, i.e., we are making mean-square-error estimations of the parameters. 

The advantage of our approach is that it does not require any mathematical derivation. Thus the mathematical complexity of the probability density function does not pose any burden for the transformer model, even though there may not be any closed form solution to MLE for the parameters for the distribution. Iterative methods (such as Newton-Raphson method or gradient descent) have been used for estimating the parameters for a distribution when there is no closed-form solution (e.g.,~\cite{r:67}). Compared to such methods, our approach has two advantages. Firstly, our approach is not iterative and just requires a single inference of a transformer model. Secondly, and more importantly, our approach does not even require knowing the probability density function of the distribution. For example, suppose a manufacturer is testing the percentage of two materials in electrical filaments in bulbs. The life of a bulb follows a distribution with a single parameter which is the percentage of the first material. Even if the distribution is unknown to us, we can still learn to estimate the parameter from many samples of bulb lives. Another example is that when testing the hyperparameters of a particular deep learning model, the prediction error on a random test case follows a distribution that is unknown to us. But we can estimate the hyperparameters based on samples of such errors.

In general, we make the following contributions in this paper:
\vspace{-0.05in}
\begin{itemize}
	\item We present a novel approach to use transformers for parameter estimation.
	\item We propose a way to convert a sample of a distribution into a sequence of embeddings, which can be easily consumed by a transformer, and can carry precise information about the sample.
	\item We conduct a comprehensive empirical study, which compares our approach with maximum likelihood estimation methods for various distributions. 
	\item It is shown that when measured by mean-square-error (MSE), our approach outperforms MLE in most commonly used distributions such as normal distribution and exponential distribution. Please note this does not indicate MLE is not a good method, as it always maximizes the probability of observing a sample. Our experiment simply indicates that our method beats MLE in terms of mean-square-error in most scenarios, which is one common way to evaluate a method of parameter estimation.
\end{itemize}

The rest of the paper is organized as follows. Related work is presented in Section 2. Section 3 describes our approach for parameter estimation with a transformer. Experimental results are reported in Section 4 and Section 5 concludes this paper.

\section*{2. Related Work}
In recent years there have been many studies that successfully applied transformer-based approaches to a diverse set of math tasks, including solving word problems~\cite{c:21}, mathematical reasoning~\cite{c:23}, symbolic regressions~\cite{c:20}, and time-series forecasting~\cite{c:20-2}. 

In contrast, there have not been many studies applying transformers to statistics. In~\cite{c:21-2} a method was proposed to convert Bayesian parameter estimation as a classification problem, which was then solved using a multi-layer perceptron network. This approach is very useful in determining if the sample was drawn from a distribution with a particular set of predetermined parameters. But it is not capable of performing parameter estimation, as it requires a hypothesis of what the parameters are. 

Although deep learning has not been used for parameter estimation in statistics, it has been used to estimate the parameters in various applications. In~\cite{r:18} the authors proposed a method to use CNN to estimate the parameters of events based on LIGO data (for gravitational waves). ~\cite{r:20} describes a method to estimate Magnetic Hamiltonian parameter from electron microscopy images using CNN. In~\cite{r:22} the authors used GRU for parameter estimation of MRI with an application in pancreatic cancer. 

Our work is different from the above in several aspects: (1) We study the problem of parameter estimation in statistics, which is key to understanding the underlying distribution of data. (2) We use transformer, which has become the state-of-the-art in most of the important applications including both NLP and computer vision. (3) Unlike the above work which uses the raw data of a specific problem as the input to their deep learning models, we convert a sample of an arbitrary distribution into a sequence of embeddings, which can be consumed by a transformer.

\section*{3. Parameter Estimation using Transformer}
\subsection*{3.1 Problem Definition}
Our goal is to train a model that can predict the parameters of a distribution, using a sample drawn from it. Let us take normal distribution as an example. A normal distribution $\mathcal{N}(\mu,\sigma^{2})$ has two parameters $\mu$ and $\sigma$, and its probability density function is p(x) = \(\frac{1}{\sigma\sqrt{2\pi}}\)$\exp(-\frac{1}{2}(\frac{x-\mu}{\sigma})^2)$. Given a sample $x_1$, …, $x_N$, the maximum likelihood estimator of $\mu$ is just the sample mean \(\frac{\sum_{i=1}^{N} x_i}{N}\). The MLE for $\sigma$ is $\sqrt{\frac{\sum_{i=1}^{N} {(x_i - \hat{\mu}})^2}{N}}$. 

We will train a model to predict the two parameters $\mu$ and $\sigma$, using a random sample drawn from the distribution as the model’s input. The model will be trained on a large number of samples, each from a different normal distribution with different parameters. In this paper we use the loss function of mean-square-error, and will evaluate the accuracy of our model by the mean-square-error of its predictions.

In this paper we only study univariate distributions, although our method can be easily extended to multivariate distributions.
\subsection*{3.2 Data Normalization}
Some distributions can be shifted and stretched along the $x$-axis (e.g., normal distribution), simply by replacing $x$ with another variable $x’ = \beta (x - \alpha)$. Therefore, given a data sample, we first normalize the range to $[0,1]$, which can be easily converted into a sequence of tokens, as described later in this section. After the model makes a prediction, the predicted parameters can be easily converted back. For example, if we replace $x$ with $x’ = \beta (x - \alpha)$ during normalization and then predict the parameters of a normal distribution to be $\mu'$ and $\sigma'$, then our predicted parameters are actually $\hat{\mu} = \frac{\mu’}{\beta} + \alpha$, and $\hat{\sigma} = \frac{\sigma’}{\beta}$.

Unlike normal distribution, some distributions can only be stretched along the $x$-axis, but cannot be shifted, such as exponential distribution. For such distributions we also normalize the range of each sample to $[0, 1]$, and again the estimated parameters can be easily converted back.

\subsection*{3.3 Data Representation}
We use a transformer with at most $L$ input embeddings, with embedding size being $K$. We use each value in each embedding to represent a possible value. For example, if $L=1024$ and $K=384$, we can represent $384K$ different values. We tried two ways to represent a value: 

\begin{itemize}
	\item \textbf{Seq-first}: First divide the range of $[0,1]$ into $L$ intervals, each corresponding to an embedding in our sequence of length $L$. Then divide each of the $L$ intervals into $K$ sub-intervals, each represented by one dimension of the embedding.  
	\item \textbf{Embed-first}: First divide the range of $[0,1]$ into $K$ intervals, each corresponding to a particular dimension in all the $L$ positions. Then divide each of the $K$ intervals into $L$ sub-intervals, to determine which position it should reside in.
\end{itemize}

Figure 1 illustrates how Seq-first conversion is done. We first divide the range $[0, 1]$ into 1024 intervals, and then each interval into 384 sub-intervals. The sample on the horizontal axis contains 6 observations. The leftmost observation is mapped to the first dimension of the first embedding. The observation 0.500001 maps to somewhere between the first and second dimensions of the 512th embedding. The first dimension of the 512th embedding maps to interval $[0.5, 0.5000025]$, and the second dimension maps to interval $[0.5000025, 0.500005]$. The observation 0.500001 falls into the interval of the first dimension. Instead of assigning all its weight to the first dimension, we assign part of its weight to the second dimension, according to its position in the first dimension’s interval, so that we can precisely represent where the observation is. In this case the observation appears at the 40-percentile position of the first interval, and thus we keep 60\% of its weight in the first dimension, and move 40\% of its weight to the second dimension. 

If we use the Embed-first representation, we assign an observation’s weight to the same dimension in two different positions, in a similar way as the Seq-first representation.

\vspace{-0.3in}
\begin{figure}
	\centering
	\includegraphics[width=0.8\linewidth]{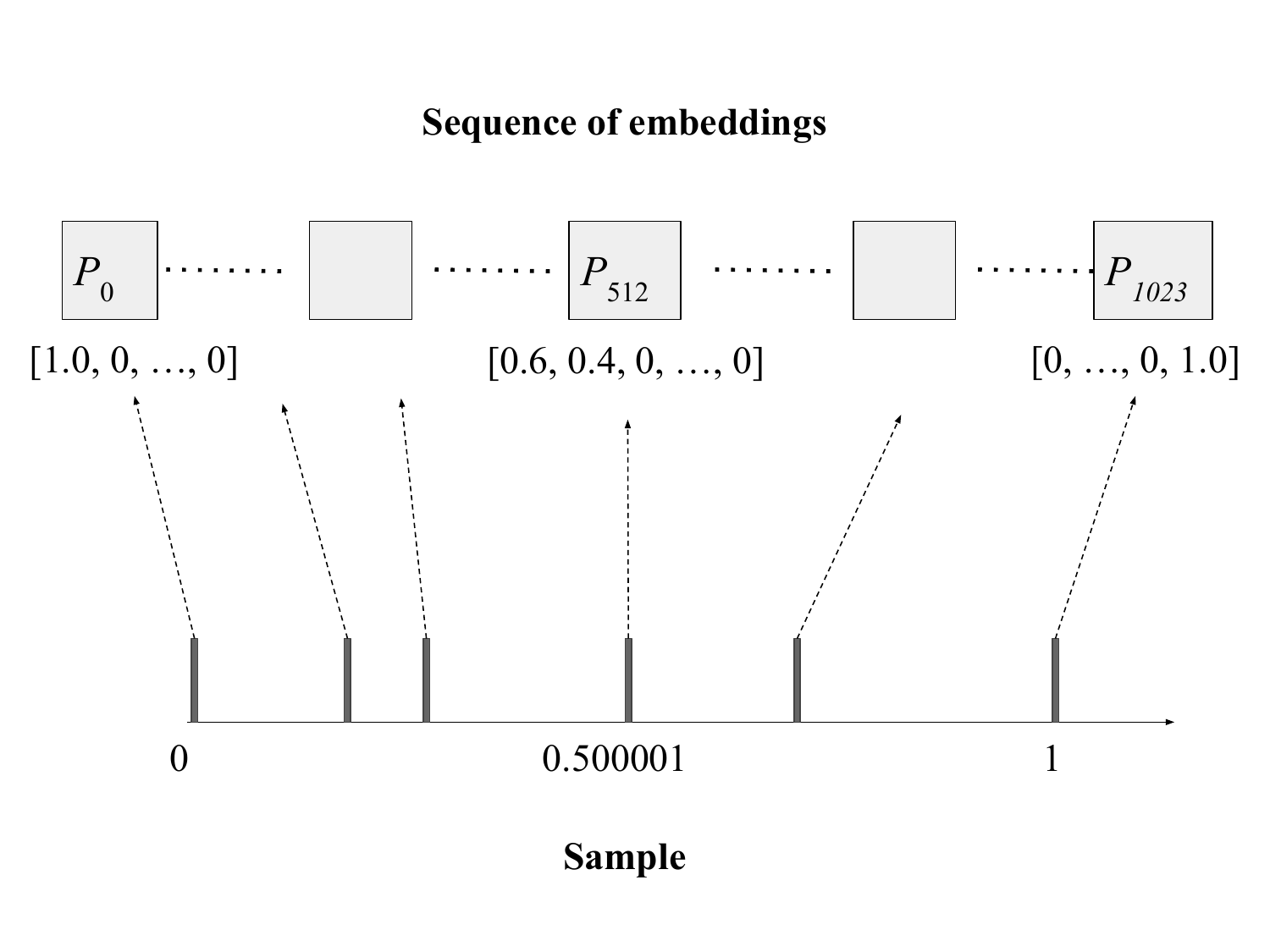}
	\vspace{-0.3in}
	\caption{Converting a sample into a sequence of $L$ embeddings ($L=1024$), each of size $K$ ($K=384$). The sample contains 6 observations, ranging from 0 to 1 (after normalization). The left-most observation is mapped to the first dimension of the first embedding. The rightmost observation is mapped to the last dimension of the last embedding. The observation 0.500001 maps to somewhere between the first and second dimensions of the 512th embedding, and thus its weight is distributed between these two dimensions. }
	\label{fig:enter-label}
	\vspace{-0.4in}
\end{figure}

\subsection*{3.4 Transformer Model}
As shown in Figure 2, our model is a typical transformer, which takes the sequence of embeddings as its input, and outputs the predicted parameter(s), by adding an output layer on top of the first output embedding of the last layer in the transformer. Since this is a regression problem, we define the loss function as the average mean-square-error on each parameter to be predicted.

\begin{figure}
	\centering
	\vspace{-0.65in}
	\includegraphics[width=0.99\linewidth]{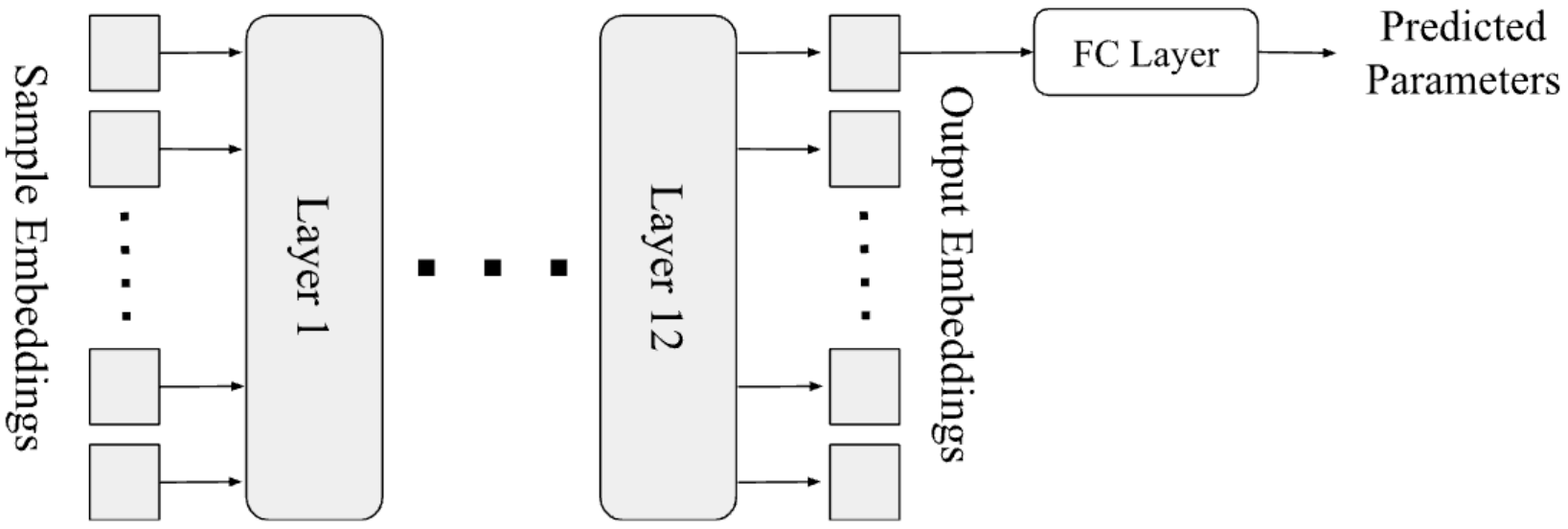}
	\vspace{-1in}
	\caption{Architecture of our transformer model}
	\label{fig:enter-label}
	\vspace{-0.2in}
\end{figure}

Each training example is a sample drawn from a distribution with random parameters. Therefore, our model should never see the same example twice during training. If the sample has been stretched or shifted as described in Section 3.2, we first convert the predicted parameters back to the original scale, and then compare them to the true parameters of the distribution for loss computation. If there are multiple parameters, we use the average mean-square-error between predicted parameter values and their true values as the loss.

\section*{4. Experiments}
\subsection*{4.1 Experiment settings}
We test our approach on three types of commonly seen distributions: Normal distributions, exponential distributions, and Poisson distributions. The training and testing data are randomly generated as needed, and thus the model should never see the same example twice. We use the Roberta model~\cite{c:19} downloaded from Huggingface as our transformer model. All experiments are done on a machine with an A6000 GPU, with Ubuntu 18.04, CUDA 11.7 and Pytorch 2.0.0. Float32 is used because numerical precision is key to our approach.

Unless otherwise mentioned, we train our model on 9.9M randomly generated examples in each setting, and test its accuracy on 100K examples. We choose our hyperparameters for efficiency reasons, and the hyperparameters chosen have negligible difference in accuracy compared with the default hyperparameters of Roberta.  The training usually takes 60 to 70 hours using the default setting. Please note that for each type of distribution we only need to train once, and 60 to 70 hours of machine time is negligible to the time consumed by a human to derive the formula or algorithm for parameter estimation.

\subsubsection*{4.1.1 Transformer Hyperparameters}
We first test various hyperparameters for our transformer model, in order to select the best settings. We start with the open-sourced RoBERTa, which accepts 512 embeddings as its input, each having 768 dimensions. In order to increase the precision, we tried to increase input length to 1024, which quadruples the memory consumption for multihead-self-attention. Due to the limitation of our GPU memory, we had to change the number of layers from 12 to 6, and embedding size from 768 to 384. As discussed in Section 3.3, we use Seq-first by default, and tried Embed-first as well. 

We use each of the above methods to estimate the parameters for normal distributions, with 9.9M samples (of size 30) for training, and 100K samples for testing. More details are described in Section 4.3.1. The results are shown in Table 1. We can see RoBERTa and RoBERTa with 1024-input-len have very similar accuracies. A $t$-test shows there is no statistically significant difference between their MSEs. We choose RoBERTa with 1024-input-len because its training time is much lower. 

\begin{table}[]
	\vspace{-0.2in}
	\begin{tabular}{|p{2.2in}|p{0.8in}|p{0.8in}|}\hline
		& MSE & Training Time\\\hline
		RoBERTa & 0.7215 & 91 hr 43 min\\\hline
		RoBERTa 1024-input-len & 0.7169 & 63 hr 35 min\\\hline
		RoBERTa 1024-input-len Embed-first & 3.155 & 63 hr 24 min\\\hline
		Maximum likelihood estimation & 0.8056 & N/A\\\hline
	\end{tabular}
	\vspace{0.1in}
	\caption{Mean-square-error and training time of various settings for normal distribution}
\end{table}

\begin{table}[]
	\vspace{-0.5in}
	\begin{tabular}{|p{0.65in}|p{1in}|p{1.5in}|p{0.6in}|p{0.6in}|}\hline
		Sample size & MLE (mean/std) & Our Approach (mean/std) & $t$-value & $p$-value\\\hline
		10 & 0.1113 / 0.1726 & 0.0796 / 0.1168 & 48.101 & 0.0001\\\hline
		30 & 0.0445 / 0.0727 & 	0.0375 / 0.0631 & 	22.995 & 0.0001\\\hline
		100 & 0.0149 / 0.0250 &	0.0143 / 0.0222 & 6.4316 & 0.0001\\\hline
		10 to 100 & 0.0503 / 0.0949 &  	0.0413 / 0.0751 & 	23.517 & 0.0001\\\hline
	\end{tabular}
	\vspace{0.1in}
	\caption{The mean-square-error of MLE and our approach for each sample size for \textbf{exponential} distribution with \textbf{known} parameter range, together two-sample $t$-test results.  }
\end{table}
\vspace{-0.5in}

\subsection*{4.2 Exponential Distribution}
We start from exponential distribution because of its simplicity. We represent the p.d.f. of exponential distribution in the same way as NumPy\footnote{https://numpy.org/doc/stable/reference/random/generated/ numpy.random.exponential.html}:

\begin{equation}
	f(x; \beta) = \frac{1}{\beta} exp (-\frac{x}{\beta})
\end{equation}

Here $\beta$ is the only parameter and represents the scale of the distribution. It is equivalent to the alternative p.d.f. $f(x;\lambda) = \lambda e^{-\lambda x}$, where $\lambda = \frac{1}{\beta}$. When generating samples, we take $\beta$ from a uniform distribution in range [0.5, 2]. 

We test our approach and maximum likelihood estimation on 100K samples, and the average mean-square-error is recorded. Each approach is tested in two settings: 
\begin{enumerate}
	\item Known Parameter Range: The range of each parameter is known to the approach.
	\item Unknown Parameter Range: The approach is not aware of the range of any parameter.
\end{enumerate}

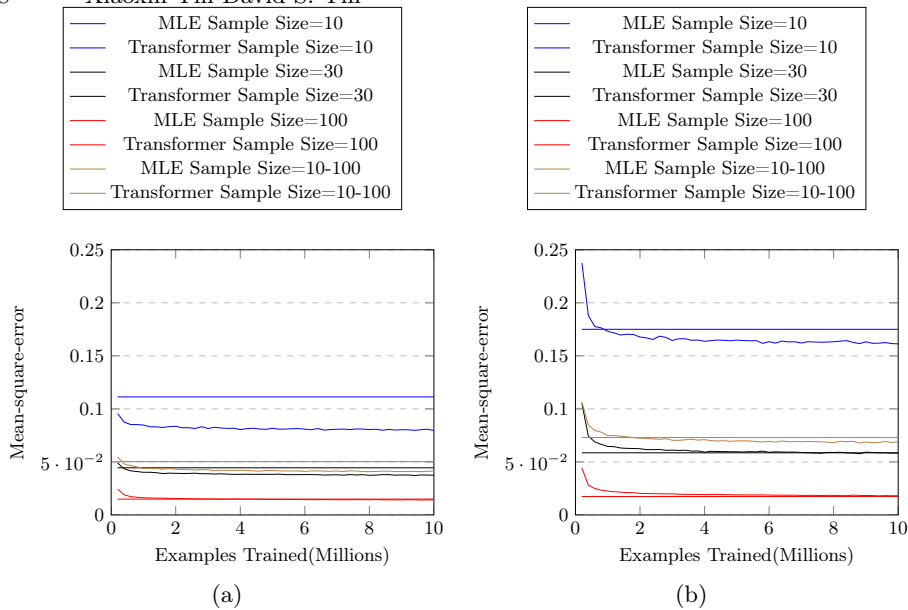
\begin{figure}
	\vspace{-0.2in}
	\begin{minipage}{.5\textwidth}
		\centering
		\resizebox{\columnwidth}{!}{
	\begin{tikzpicture}
		\begin{axis}[
			title={},
			xlabel={Examples Trained(Millions)},
			ylabel={Mean-square-error},
			xmin=0, xmax=10,
			ymin=0, ymax=0.25,
			xtick={0,2,4,6,8,10},
			ytick={0,0.05,0.1,0.15,0.2,0.25},
			legend style={at={(-0.15,1.15)},anchor=south west},
			ymajorgrids=true,
			grid style=dashed,
			]
			\addplot[
			color=blue,
			mark=circle,
			]
			coordinates {
				(0.2,0.11133150188133964)
				(10,0.11133150188133964)
			};
			\addlegendentry{MLE Sample Size=10}
			\addplot[
			color=blue,
			mark=circle,
			]
			coordinates {
				(0.2, 0.0955771)
				(0.4, 0.087548471)
				(0.6, 0.085281353)
				(0.8, 0.085177479)
				(1, 0.0847534759999999)
				(1.2, 0.0831471389999999)
				(1.4, 0.0830772949999999)
				(1.6, 0.08251258)
				(1.8, 0.083263392)
				(2, 0.08360906)
				(2.2, 0.0821871789999999)
				(2.4, 0.0822809659999999)
				(2.6, 0.081472945)
				(2.8, 0.0832262849999999)
				(3, 0.081530535)
				(3.2, 0.082660908)
				(3.4, 0.0819727479999999)
				(3.6, 0.081616987)
				(3.8, 0.081749012)
				(4, 0.0805292249999999)
				(4.2, 0.08135011)
				(4.4, 0.0817626059999999)
				(4.6, 0.081005681)
				(4.8, 0.080773028)
				(5, 0.08172104)
				(5.2, 0.080849452)
				(5.4, 0.080989444)
				(5.6, 0.0813023179999999)
				(5.8, 0.081416028)
				(6, 0.080931003)
				(6.2, 0.0803279609999999)
				(6.4, 0.081485727)
				(6.6, 0.079996413)
				(6.8, 0.080535175)
				(7, 0.080800751)
				(7.2, 0.080946346)
				(7.4, 0.080539449)
				(7.6, 0.0810393629999999)
				(7.8, 0.08097807)
				(8, 0.080484294)
				(8.2, 0.080794437)
				(8.4, 0.079897813)
				(8.6, 0.080658617)
				(8.8, 0.080072273)
				(9, 0.080312292)
				(9.2, 0.080311665)
				(9.4, 0.079788354)
				(9.6, 0.080384581)
				(9.8, 0.0808023)
				(10, 0.079638552)
			};
			\addlegendentry{Transformer Sample Size=10}
			\addplot[
			color=black,
			mark=circle,
			]
			coordinates {
				(0.2,0.04446472274014219)
				(10,0.04446472274014219)
			};
			\addlegendentry{MLE Sample Size=30}
			\addplot[
			color=black,
			mark=circle,
			]
			coordinates {
				(0.2, 0.048994799)
				(0.4, 0.043525138)
				(0.6, 0.041610858)
				(0.8, 0.040905603)
				(1, 0.0402389829999999)
				(1.2, 0.0402449339999999)
				(1.4, 0.039843411)
				(1.6, 0.0390442089999999)
				(1.8, 0.0394181139999999)
				(2, 0.039253383)
				(2.2, 0.039129784)
				(2.4, 0.039141057)
				(2.6, 0.039366218)
				(2.8, 0.0388949499999999)
				(3, 0.038848999)
				(3.2, 0.038580354)
				(3.4, 0.03881781)
				(3.6, 0.038469634)
				(3.8, 0.038303232)
				(4, 0.038300765)
				(4.2, 0.038389958)
				(4.4, 0.038290784)
				(4.6, 0.0383338689999999)
				(4.8, 0.038297407)
				(5, 0.038303245)
				(5.2, 0.0384832099999999)
				(5.4, 0.038004719)
				(5.6, 0.038004622)
				(5.8, 0.0380307629999999)
				(6, 0.038059623)
				(6.2, 0.038096247)
				(6.4, 0.037938695)
				(6.6, 0.0376682)
				(6.8, 0.0377581609999999)
				(7, 0.0378743239999999)
				(7.2, 0.03756441)
				(7.4, 0.038053004)
				(7.6, 0.0377072689999999)
				(7.8, 0.037938496)
				(8, 0.037399134)
				(8.2, 0.037562282)
				(8.4, 0.0380122909999999)
				(8.6, 0.037985957)
				(8.8, 0.037536779)
				(9, 0.037660076)
				(9.2, 0.037676631)
				(9.4, 0.0372218829999999)
				(9.6, 0.037677607)
				(10, 0.037512878)
			};
			\addlegendentry{Transformer Sample Size=30}   
			\addplot[
			color=red,
			mark=circle,
			]
			coordinates {
				(0.2,0.014944702481521029)
				(10,0.014944702481521029)
			};
			\addlegendentry{MLE Sample Size=100}
			\addplot[
			color=red,
			mark=circle,
			]
			coordinates {
				(0.2, 0.024494234)
				(0.4, 0.019050254)
				(0.6, 0.017292692)
				(0.8, 0.016660355)
				(1, 0.0162221909999999)
				(1.2, 0.016014439)
				(1.4, 0.015873882)
				(1.6, 0.015849773)
				(1.8, 0.015675859)
				(2, 0.015536492)
				(2.2, 0.01555929)
				(2.4, 0.015296142)
				(2.6, 0.015194717)
				(2.8, 0.015065845)
				(3, 0.0150487149999999)
				(3.2, 0.014983899)
				(3.4, 0.0148950239999999)
				(3.6, 0.0148640949999999)
				(3.8, 0.015033003)
				(4, 0.014952468)
				(4.2, 0.014911979)
				(4.4, 0.014871868)
				(4.6, 0.014845288)
				(4.8, 0.0146471359999999)
				(5, 0.014669741)
				(5.2, 0.014631522)
				(5.4, 0.014710497)
				(5.6, 0.014596301)
				(5.8, 0.014574165)
				(6, 0.014456068)
				(6.2, 0.014413992)
				(6.4, 0.0144608189999999)
				(6.6, 0.0145137509999999)
				(6.8, 0.014411215)
				(7, 0.014374801)
				(7.2, 0.014306632)
				(7.4, 0.014370982)
				(7.6, 0.0144584009999999)
				(7.8, 0.01441901)
				(8, 0.0143032089999999)
				(8.2, 0.014466235)
				(8.4, 0.0143073459999999)
				(8.6, 0.014231639)
				(8.8, 0.014252619)
				(9, 0.014195776)
				(9.2, 0.014317109)
				(9.4, 0.01433527)
				(9.6, 0.014083945)
				(9.8, 0.014238016)
				(10, 0.01426282)
			};
			\addlegendentry{Transformer Sample Size=100}
			\addplot[
			color=brown,
			mark=circle,
			]
			coordinates {
				(0.2,0.05029182354363478)
				(10,0.05029182354363478)
			};
			\addlegendentry{MLE Sample Size=10-100}
			\addplot[
			color=brown,
			mark=circle,
			]
			coordinates {
				(0.2, 0.054738385)
				(0.4, 0.0480939459999999)
				(0.6, 0.046557514)
				(0.8, 0.045240385)
				(1, 0.043875216)
				(1.2, 0.04398655)
				(1.4, 0.043682968)
				(1.6, 0.043202677)
				(1.8, 0.043729416)
				(2, 0.043186296)
				(2.2, 0.042808396)
				(2.4, 0.0427208699999999)
				(2.6, 0.042370902)
				(2.8, 0.042730105)
				(3, 0.042639041)
				(3.2, 0.041693514)
				(3.4, 0.0421812759999999)
				(3.6, 0.041985362)
				(3.8, 0.041978449)
				(4, 0.042058392)
				(4.2, 0.04200135)
				(4.4, 0.04156012)
				(4.6, 0.0412529859999999)
				(4.8, 0.0419250419999999)
				(5, 0.041934941)
				(5.2, 0.041593186)
				(5.4, 0.041465126)
				(5.6, 0.0414539869999999)
				(5.8, 0.0414568959999999)
				(6, 0.041030832)
				(6.2, 0.0415337509999999)
				(6.4, 0.041559209)
				(6.6, 0.041307767)
				(6.8, 0.041173384)
				(7, 0.0414416159999999)
				(7.2, 0.040981844)
				(7.4, 0.041535892)
				(7.6, 0.041414268)
				(7.8, 0.0409035489999999)
				(8, 0.040780652)
				(8.2, 0.041028709)
				(8.4, 0.041296123)
				(8.6, 0.041299596)
				(8.8, 0.0408617679999999)
				(9, 0.040909905)
				(9.2, 0.041077765)
				(9.4, 0.0410512649999999)
				(9.6, 0.040793427)
				(9.8, 0.041180638)
				(10, 0.0412939809999999)
			};
			\addlegendentry{Transformer Sample Size=10-100}
		\end{axis}
	\end{tikzpicture}
}
(a)
\end{minipage}
\begin{minipage}{.5\textwidth}
	\centering
	\resizebox{\columnwidth}{!}{
	\begin{tikzpicture}
		\begin{axis}[
			title={},
			xlabel={Examples Trained(Millions)},
			ylabel={Mean-square-error},
			xmin=0, xmax=10,
			ymin=0, ymax=0.25,
			xtick={0,2,4,6,8,10},
			ytick={0,0.05,0.1,0.15,0.2,0.25},
			legend style={at={(-0.15,1.15)},anchor=south west},
			ymajorgrids=true,
			grid style=dashed,
			]
			\addplot[
			color=blue,
			mark=circle,
			]
			coordinates {
				(0.2, 0.17503596605934352)
				(10, 0.17503596605934352)
			};
			\addlegendentry{MLE Sample Size=10}
			\addplot[
			color=blue,
			mark=circle,
			]
			coordinates {
				(0.2, 0.237771708)
				(0.4, 0.187915203)
				(0.6, 0.177635482)
				(0.8, 0.176332643999999)
				(1, 0.173141043)
				(1.2, 0.171609421)
				(1.4, 0.169682746)
				(1.6, 0.170361252)
				(1.8, 0.170070060999999)
				(2, 0.167770366)
				(2.2, 0.166952823)
				(2.4, 0.165408349)
				(2.6, 0.168552866)
				(2.8, 0.167412863)
				(3, 0.164512229)
				(3.2, 0.166107765999999)
				(3.4, 0.166136256)
				(3.6, 0.164701158)
				(3.8, 0.164916766)
				(4, 0.163674063999999)
				(4.2, 0.164354073)
				(4.4, 0.164898981)
				(4.6, 0.164653732999999)
				(4.8, 0.164196286)
				(5, 0.164889047999999)
				(5.2, 0.16443997)
				(5.4, 0.164214527)
				(5.6, 0.164324554999999)
				(5.8, 0.161816792)
				(6, 0.163358609)
				(6.2, 0.162011219999999)
				(6.4, 0.163955592)
				(6.6, 0.16311925)
				(6.8, 0.163297641)
				(7, 0.162113985999999)
				(7.2, 0.163356702)
				(7.4, 0.16307178)
				(7.6, 0.162841754)
				(7.8, 0.163023685)
				(8, 0.163288752)
				(8.2, 0.163849561)
				(8.4, 0.164361139999999)
				(8.6, 0.162739928999999)
				(8.8, 0.161557339)
				(9, 0.163251606)
				(9.2, 0.161855570999999)
				(9.4, 0.161941244999999)
				(9.6, 0.162524857)
				(9.8, 0.161564359)
				(10, 0.161297154)
			};
			\addlegendentry{Transformer Sample Size=10}
			\addplot[
			color=black,
			mark=circle,
			]
			coordinates {
				(0.2,0.058594892186480084)
				(10,0.058594892186480084)
			};
			\addlegendentry{MLE Sample Size=30}
			\addplot[
			color=black,
			mark=circle,
			]
			coordinates {
				(0.2, 0.105200489)
				(0.4, 0.074310281)
				(0.6, 0.0688674959999999)
				(0.8, 0.0665589639999999)
				(1, 0.064601932)
				(1.2, 0.064378118)
				(1.4, 0.063137502)
				(1.6, 0.062760596)
				(1.8, 0.062669085)
				(2, 0.062428301)
				(2.2, 0.061614504)
				(2.4, 0.06159556)
				(2.6, 0.0617728319999999)
				(2.8, 0.0609493449999999)
				(3, 0.061113868)
				(3.2, 0.0608489539999999)
				(3.4, 0.0610413049999999)
				(3.6, 0.060175997)
				(3.8, 0.06034009)
				(4, 0.059214688)
				(4.2, 0.06001559)
				(4.4, 0.059792851)
				(4.6, 0.059742266)
				(4.8, 0.0595955719999999)
				(5, 0.059593936)
				(5.2, 0.059312215)
				(5.4, 0.059727807)
				(5.6, 0.059461551)
				(5.8, 0.0601689779999999)
				(6, 0.059067961)
				(6.2, 0.059521592)
				(6.4, 0.059328486)
				(6.6, 0.0592222239999999)
				(6.8, 0.059260388)
				(7, 0.058861017)
				(7.2, 0.0586749709999999)
				(7.4, 0.0584180319999999)
				(7.6, 0.058216181)
				(7.8, 0.058986808)
				(8, 0.0589007639999999)
				(8.2, 0.0590953749999999)
				(8.4, 0.0585540959999999)
				(8.6, 0.058389272)
				(8.8, 0.057693105)
				(9, 0.059224918)
				(9.2, 0.058858051)
				(9.4, 0.05858718)
				(9.6, 0.058610717)
				(9.8, 0.05827365)
				(10, 0.058373915)
			};
			\addlegendentry{Transformer Sample Size=30}   
			\addplot[
			color=red,
			mark=circle,
			]
			coordinates {
				(0.2,0.017418725401989126)
				(10,0.017418725401989126)
			};
			\addlegendentry{MLE Sample Size=100}
			\addplot[
			color=red,
			mark=circle,
			]
			coordinates {
				(0.2, 0.0441644509999999)
				(0.4, 0.0280454289999999)
				(0.6, 0.024965858)
				(0.8, 0.0233882369999999)
				(1, 0.0224712529999999)
				(1.2, 0.02195221)
				(1.4, 0.02145493)
				(1.6, 0.021098096)
				(1.8, 0.0210300309999999)
				(2, 0.020305247)
				(2.2, 0.020233127)
				(2.4, 0.020071027)
				(2.6, 0.019990636)
				(2.8, 0.019879011)
				(3, 0.019875454)
				(3.2, 0.0195974769999999)
				(3.4, 0.019440681)
				(3.6, 0.019396012)
				(3.8, 0.0192960259999999)
				(4, 0.01916138)
				(4.2, 0.019347272)
				(4.4, 0.0194019169999999)
				(4.6, 0.019140314)
				(4.8, 0.0191210419999999)
				(5, 0.0189668339999999)
				(5.2, 0.018929325)
				(5.4, 0.018932357)
				(5.6, 0.018867622)
				(5.8, 0.0189492039999999)
				(6, 0.0187737539999999)
				(6.2, 0.018649262)
				(6.4, 0.018656678)
				(6.6, 0.018582671)
				(6.8, 0.0186628229999999)
				(7, 0.018396258)
				(7.2, 0.018468609)
				(7.4, 0.018517921)
				(7.6, 0.018287074)
				(7.8, 0.018486912)
				(8, 0.018447899)
				(8.2, 0.018179377)
				(8.4, 0.018230011)
				(8.6, 0.018634503)
				(8.8, 0.0184476489999999)
				(9, 0.018317417)
				(9.2, 0.018303918)
				(9.4, 0.01803207)
				(9.6, 0.018254128)
				(9.8, 0.018087336)
				(10, 0.018167312)
			};
			\addlegendentry{Transformer Sample Size=100}
			\addplot[
			color=brown,
			mark=circle,
			]
			coordinates {
				(0.2,0.07298855502898413)
				(10,0.07298855502898413)
			};
			\addlegendentry{MLE Sample Size=10-100}
			\addplot[
			color=brown,
			mark=circle,
			]
			coordinates {
				(0.2, 0.106488245)
				(0.4, 0.0845369219999999)
				(0.6, 0.07970132)
				(0.8, 0.07807184)
				(1, 0.074689806)
				(1.2, 0.074875882)
				(1.4, 0.074239752)
				(1.6, 0.073706115)
				(1.8, 0.072825465)
				(2, 0.072402054)
				(2.2, 0.0722393179999999)
				(2.4, 0.07174026)
				(2.6, 0.0712836439999999)
				(2.8, 0.071885774)
				(3, 0.07041033)
				(3.2, 0.0703252549999999)
				(3.4, 0.0708892589999999)
				(3.6, 0.071384985)
				(3.8, 0.07056014)
				(4, 0.0709758399999999)
				(4.2, 0.070090972)
				(4.4, 0.069700227)
				(4.6, 0.0700306199999999)
				(4.8, 0.069411933)
				(5, 0.069827983)
				(5.2, 0.069856833)
				(5.4, 0.069780554)
				(5.6, 0.069165421)
				(5.8, 0.0688883399999999)
				(6, 0.069388602)
				(6.2, 0.068762609)
				(6.4, 0.069724948)
				(6.6, 0.068996776)
				(6.8, 0.069247997)
				(7, 0.069062944)
				(7.2, 0.069278369)
				(7.4, 0.069134736)
				(7.6, 0.068973291)
				(7.8, 0.06868179)
				(8, 0.06814688)
				(8.2, 0.069460974)
				(8.4, 0.069079834)
				(8.6, 0.0694576879999999)
				(8.8, 0.067794406)
				(9, 0.0687339859999999)
				(9.2, 0.068065464)
				(9.4, 0.068408354)
				(9.6, 0.069286025)
				(9.8, 0.06829088)
				(10, 0.069121017)
			};
			\addlegendentry{Transformer Sample Size=10-100}
		\end{axis}
	\end{tikzpicture}
}
(b)
	\end{minipage} 
	\caption{(a) The mean-square-error with \# training examples for \textbf{exponential} distributions with \textbf{known} parameter ranges. The horizontal lines represent mean-square-errors of MLE, and the curves represent those of our approach. (b) Those for \textbf{exponential} distributions with \textbf{unknown} parameter ranges.}
	\vspace{-0.2in}
\end{figure}

\subsubsection*{4.2.1 Known Parameter Range}
When each parameter’s range is known, we normalize each sample into [0, 1] by a fixed linear transform. When $\beta \in [0.5, 2]$, the probability of $x > 20$ is less than 0.5\%. Therefore, we cap each value of a sample within [0, 20], and normalize each capped value $x_i$ by $x_i’ = x_i / 20$. In the final fully-connected layer for predicting parameters (as in Figure 2), we let it directly predict the value of the parameter. 

The MLE of $\beta$ is simply $\frac{\sum_{i=1}^{N} x_i}{N}$, and we cap its estimate of $\beta$ within [0.5, 2]. 

We test with three different sample sizes (10, 30, and 100), and another setting in which the sample size is randomly sampled from a log-uniform distribution of range [10, 100]. The results are shown in Figure 3 (a), in which our method outperforms MLE for each sample size. Table 2 shows the mean and standard deviation of the mean-square-error of the two approaches for each sample size. One can see that our proposed approach solidly beats MLE in every sample size, with close to zero $p$-values in two-sample $t$-tests.

\subsubsection*{4.2.2 Unknown Parameter Range}

Suppose a sample is drawn from an exponential distribution $exp(\beta)$, and the range of $\beta$ is unknown to the method for parameter estimation (i.e., the method needs to assume that $\beta$ can take any value). For MLE, we can simply use its formula (as in Section 4.2.1) to estimate the parameters. But a slightly more complex method is needed for our approach.

In order to hide the parameter’s range from our transformer model, we first normalize each sample $s$ into range $[0, 1]$. Let $b = max(s)$. Each value $x_i$ is normalized by $x_i’ = x_i / b$. In this way our model can only see the relative shape of the sample, without knowing its original range. Suppose the model’s prediction is $\beta^{*}$. To compute the loss, we first recover the estimated parameter into the original range: The estimate of $\beta$ is $\hat{\beta} = \beta^{*} \cdot b$. Then we compare it with the true parameter, as follows:

\vspace{-0.1in}
\begin{equation}
	loss = \mbox{mean-square-error}(\hat{\beta}, \beta)
\end{equation}

Please note that $b$ is never seen by the model, and thus the model is unaware of the range of the sample or the parameter. The loss function is just the mean-square-error between the estimated parameters and true parameters, which is what we are optimizing.

Again we test with three different sample sizes (10, 30, and 100), and random sample size from range [10, 100]. The results are shown in Figure 3(b) and Table 3. We can see that our method outperforms MLE in most cases, except when the sample size is large.

\begin{table}[]
	\vspace{-0.2in}
	\begin{tabular}{|p{0.65in}|p{1in}|p{1.5in}|p{0.6in}|p{0.6in}|}\hline
		Sample size & MLE (mean/std) & Our Approach (mean/std) & $t$-value & $p$-value\\\hline
		10 & 0.1750 / 0.3479 & 0.1613 / 0.2816 & 9.6793 & 0.0001\\\hline
		30 & 0.0586 / 0.1078 & 0.0584 / 0.1001 & 0.4299 & 0.6673\\\hline
		100 & 0.0174 / 0.0312 &	0.0182 / 0.0323 & -5.2726 & 0.0001\\\hline
		10 to 100 & 0.0730 / 0.1763 & 0.0691 / 0.1468 & 5.3758 & 0.0001\\\hline
	\end{tabular}
	\vspace{0.1in}
	\caption{The mean-square-error of MLE and our approach for each sample size for \textbf{exponential} distribution with \textbf{unknown} parameter range, together two-sample $t$-test results. }
	\vspace{-0.4in}
\end{table}

\subsection*{4.3 Normal Distribution}
We then test our approach on normal distribution and compare with maximum likelihood estimation. The parameter $\mu$ is drawn from a uniform distribution with range $[-5, 5]$, and $\sigma$ from that with range $[1, 10]$, independently from $\mu$. 

\subsubsection*{4.3.1 Known Parameter Range}
When each parameter’s range is known, our approach will normalize each sample into $[0, 1]$ by a fixed linear transform. In this case it caps each value of a sample within $[-35, 35]$ (because 35 is at least three standard deviations away from the mean), and normalizes each capped value $x_i$ by $x_i’ = ( x_i + 35 ) / 70$. 

For maximum likelihood estimation, the MLE of $\mu$ is $\hat{\mu} = \frac{\sum_{i=1}^{N} x_i}{N}$, and that of $\sigma$ is $\hat{\sigma} = \sqrt{\frac{\sum_{i=1}^{N}{(x_i - \hat{\mu})^2}}{N}}$. We cap $\hat{\mu}$ and $\hat{\sigma}$ within their ranges ($[-5, 5]$ and $[1, 10]$) when the estimates are out of that range.

\begin{figure}
	\begin{minipage}{.5\textwidth}
		\centering
		\resizebox{\columnwidth}{!}{
	\begin{tikzpicture}
		\begin{axis}[
			title={},
			xlabel={Examples Trained (Millions)},
			ylabel={Mean-square-error},
			xmin=0, xmax=10,
			ymin=0, ymax=2.5,
			xtick={0,2,4,6,8,10},
			ytick={0,0.5,1,1.5,2,2.5},
			legend style={at={(-0.15,1.15)},anchor=south west},
			ymajorgrids=true,
			grid style=dashed,
			]
			\addplot[
			color=blue,
			mark=circle,
			]
			coordinates {
				(0.2,2.2809454734)
				(10,2.2809454734)
			};
			\addlegendentry{MLE Sample Size=10}
			\addplot[
			color=blue,
			mark=circle,
			]
			coordinates {
				(0.2, 2.30362976)
				(0.4, 1.89160595999999)
				(0.6, 1.83489381999999)
				(0.8, 1.81178174)
				(1, 1.81253842)
				(1.2, 1.80117083999999)
				(1.4, 1.78081554)
				(1.6, 1.77104374)
				(1.8, 1.79817006)
				(2, 1.7704987)
				(2.2, 1.77926448)
				(2.4, 1.77084681999999)
				(2.6, 1.76312842)
				(2.8, 1.77858247999999)
				(3, 1.76215696)
				(3.2, 1.7699822)
				(3.4, 1.75896508)
				(3.6, 1.74843808)
				(3.8, 1.75645435999999)
				(4, 1.7562037)
				(4.2, 1.75312352)
				(4.4, 1.74952932)
				(4.6, 1.74414106)
				(4.8, 1.75848456)
				(5, 1.76576997999999)
				(5.2, 1.73804024)
				(5.4, 1.75791035999999)
				(5.6, 1.73692784)
				(5.8, 1.75483576)
				(6, 1.74229359999999)
				(6.2, 1.73986541999999)
				(6.4, 1.75857698)
				(6.6, 1.74683788)
				(6.8, 1.74846748)
				(7, 1.74765366)
				(7.2, 1.74274185999999)
				(7.4, 1.747754)
				(7.6, 1.75650014)
				(7.8, 1.7402787)
				(8, 1.72708216)
				(8.2, 1.75108974)
				(8.4, 1.73923501999999)
				(8.6, 1.76093075999999)
				(8.8, 1.73794158)
				(9, 1.7448905)
				(9.2, 1.72120137999999)
				(9.4, 1.7245549)
				(9.6, 1.72931299999999)
				(9.8, 1.72846973999999)
				(10, 1.73372976)
			};
			\addlegendentry{Transformer Sample Size=10}
			\addplot[
			color=black,
			mark=circle,
			]
			coordinates {
				(0.2,0.80559629997)
				(10,0.80559629997)
			};
			\addlegendentry{MLE Sample Size=30}
			\addplot[
			color=black,
			mark=circle,
			]
			coordinates {
				(0.2, 1.00267796)
				(0.4, 0.815175824999999)
				(0.6, 0.787378954999999)
				(0.8, 0.775531389999999)
				(1, 0.763443244999999)
				(1.2, 0.75341956)
				(1.4, 0.749440174999999)
				(1.6, 0.7552115)
				(1.8, 0.742528085)
				(2, 0.744664945)
				(2.2, 0.74622819)
				(2.4, 0.747792405)
				(2.6, 0.745279055)
				(2.8, 0.74063575)
				(3, 0.74482454)
				(3.2, 0.73580588)
				(3.4, 0.7400678)
				(3.6, 0.73254222)
				(3.8, 0.74416678)
				(4, 0.739323864999999)
				(4.2, 0.741905299999999)
				(4.4, 0.730391005)
				(4.6, 0.736037684999999)
				(4.8, 0.732529189999999)
				(5, 0.737953799999999)
				(5.2, 0.729841714999999)
				(5.4, 0.73403244)
				(5.6, 0.741853515)
				(5.8, 0.733041495)
				(6, 0.72857072)
				(6.2, 0.737060305)
				(6.4, 0.731552284999999)
				(6.6, 0.733584369999999)
				(6.8, 0.733795724999999)
				(7, 0.731444354999999)
				(7.2, 0.721823815)
				(7.4, 0.719530819999999)
				(7.6, 0.73048726)
				(7.8, 0.71816849)
				(8, 0.72440854)
				(8.2, 0.732424909999999)
				(8.4, 0.727388754999999)
				(8.6, 0.72313789)
				(8.8, 0.72555106)
				(9, 0.718245415)
				(9.2, 0.723186489999999)
				(9.4, 0.722250084999999)
				(9.6, 0.72443309)
				(9.8, 0.7165661)
				(10, 0.716920005)
			};
			\addlegendentry{Transformer Sample Size=30}   
			\addplot[
			color=red,
			mark=circle,
			]
			coordinates {
				(0.2,0.25585851061)
				(10,0.25585851061)
			};
			\addlegendentry{MLE Sample Size=100}
			\addplot[
			color=red,
			mark=circle,
			]
			coordinates {
				(0.2, 0.432517676)
				(0.4, 0.323972706)
				(0.6, 0.29804728)
				(0.8, 0.287901869)
				(1,0.281839471)
				(1.2, 0.275388635)
				(1.4, 0.273018812)
				(1.6, 0.27011606)
				(1.8, 0.270968146)
				(2, 0.26588931)
				(2.2, 0.263885539)
				(2.4, 0.26469579)
				(2.6, 0.262041726999999)
				(2.8, 0.262446291999999)
				(3, 0.264115895999999)
				(3.2, 0.259859892)
				(3.4, 0.260647651999999)
				(3.6, 0.261932778)
				(3.8, 0.261247411)
				(4, 0.259894022)
				(4.2, 0.258451941)
				(4.4, 0.25947139)
				(4.6, 0.257232826)
				(4.8, 0.25487001)
				(5, 0.258932717)
				(5.2, 0.257996271)
				(5.4, 0.257146859999999)
				(5.6, 0.25527302)
				(5.8, 0.254203111)
				(6, 0.255397871)
				(6.2, 0.255935349)
				(6.4, 0.252998530999999)
				(6.6, 0.255279811999999)
				(6.8, 0.252970463)
				(7, 0.253873132)
				(7.2, 0.253029144)
				(7.4, 0.252496265999999)
				(7.6, 0.252020803999999)
				(7.8, 0.252551682)
				(8, 0.251299576)
				(8.2, 0.253764607)
				(8.4, 0.249686864)
				(8.6, 0.251101379)
				(8.8, 0.251323822)
				(9, 0.251320951)
				(9.2, 0.24948301)
				(9.4, 0.250238313)
				(9.6, 0.251097605)
				(9.8, 0.247831564)
				(10, 0.249203401)
			};
			\addlegendentry{Transformer Sample Size=100}
			\addplot[
			color=brown,
			mark=circle,
			]
			coordinates {
				(0.2,0.9636363178)
				(10,0.9636363178)
			};
			\addlegendentry{MLE Sample Size=10-100}
			\addplot[
			color=brown,
			mark=circle,
			]
			coordinates {
				(0.2, 1.22363440999999)
				(0.4, 0.9739831)
				(0.6, 0.91316224)
				(0.8, 0.8829605)
				(1, 0.86793947)
				(1.2, 0.87123519)
				(1.4, 0.854330649999999)
				(1.6, 0.84808309)
				(1.8, 0.847302914999999)
				(2, 0.847072925)
				(2.2, 0.844190745)
				(2.4, 0.8430025)
				(2.6, 0.843950364999999)
				(2.8, 0.841937115)
				(3, 0.833510259999999)
				(3.2, 0.83487471)
				(3.4, 0.83686121)
				(3.6, 0.83360443)
				(3.8, 0.83887851)
				(4, 0.829386055)
				(4.2, 0.82606364)
				(4.4, 0.825597805)
				(4.6, 0.82657567)
				(4.8, 0.83064656)
				(5, 0.8235629)
				(5.2, 0.820132335)
				(5.4, 0.82357669)
				(5.6, 0.818281375)
				(5.8, 0.815942084999999)
				(6, 0.830523185)
				(6.2, 0.82733995)
				(6.4, 0.828067035)
				(6.6, 0.809995095)
				(6.8, 0.81424064)
				(7, 0.822084855)
				(7.2, 0.821133545)
				(7.4, 0.81743389)
				(7.6, 0.82264408)
				(7.8, 0.80416821)
				(8, 0.813349335)
				(8.2, 0.809359785)
				(8.4, 0.823865735)
				(8.6, 0.82205523)
				(8.8, 0.818808575)
				(9, 0.816938945)
				(9.2, 0.813310905)
				(9.4, 0.810289104999999)
				(9.6, 0.820063355)
				(9.8, 0.807292179999999)
				(10, 0.810590275)
			};
			\addlegendentry{Transformer Sample Size=10-100}
		\end{axis}
	\end{tikzpicture}
}
(a)
\end{minipage}
\begin{minipage}{.5\textwidth}
	\centering
	\resizebox{\columnwidth}{!}{
		\begin{tikzpicture}
			\begin{axis}[
				title={},
				xlabel={Examples Trained(Millions)},
				ylabel={Mean-square-error},
				xmin=0, xmax=10,
				ymin=0, ymax=4,
				xtick={0,2,4,6,8,10},
				ytick={0,0.5,1,1.5,2,2.5,3,3.5,4},
				legend style={at={(0,1.15)},anchor=south west},
				ymajorgrids=true,
				grid style=dashed,
				]
				\addplot[
				color=blue,
				mark=circle,
				]
				coordinates {
					(0.2, 3.42891544)
					(0.4, 3.12984374)
					(0.6, 3.07958591999999)
					(0.8, 3.03301289999999)
					(1, 2.98188254)
					(1.2, 2.98306997999999)
					(1.4, 2.97345986)
					(1.6, 2.9671426)
					(1.8, 2.94895802)
					(2, 2.93572147999999)
					(2.2, 2.89401164)
					(2.4, 2.91227836)
					(2.6, 2.92498352)
					(2.8, 2.91614915999999)
					(3, 2.89075654)
					(3.2, 2.89632444)
					(3.4, 2.88787084)
					(3.6, 2.93206476)
					(3.8, 2.94140354)
					(4, 2.90714354)
					(4.2, 2.91989028)
					(4.4, 2.90431724)
					(4.6, 2.90307464)
					(4.8, 2.88899079999999)
					(5, 2.89339402)
					(5.2, 2.89628891999999)
					(5.4, 2.89691528)
					(5.6, 2.88980896)
					(5.8, 2.91260512)
					(6, 2.88450532)
					(6.2, 2.87244019999999)
					(6.4, 2.85298569999999)
					(6.6, 2.87386077999999)
					(6.8, 2.86204351999999)
					(7, 2.85714354)
					(7.2, 2.85161512)
					(7.4, 2.8709068)
					(7.6, 2.89187932)
					(7.8, 2.90219116)
					(8, 2.86515162)
					(8.2, 2.87045562)
					(8.4, 2.89286354)
					(8.6, 2.86258288)
					(8.8, 2.85564451999999)
					(9, 2.88384826)
					(9.2, 2.86291084)
					(9.4, 2.85444202)
					(9.6, 2.8599481)
					(9.8, 2.8730135)
					(10, 2.87684961999999)
				};
				\addlegendentry{Transformer Sample Size=10}
				\addplot[
				color=blue,
				mark=circle,
				]
				coordinates {
					(0.2, 2.8637471599080135)
					(10, 2.8637471599080135)
				};
				\addlegendentry{MLE Sample Size=10}
				\addplot[
				color=black,
				mark=circle,
				]
				coordinates {
					(0.2, 1.35780812)
					(0.4, 1.15625924)
					(0.6, 1.07929402)
					(0.8, 1.07278115999999)
					(1, 1.04519212999999)
					(1.2, 1.03403428)
					(1.4, 1.02153597999999)
					(1.6, 1.01965366999999)
					(1.8, 1.00738333)
					(2, 1.00063219)
					(2.2, 0.99594479)
					(2.4, 0.98546296)
					(2.6, 0.99421165)
					(2.8, 0.98938467)
					(3, 0.98712268)
					(3.2, 0.975567919999999)
					(3.4, 0.97856946)
					(3.6, 0.986263339999999)
					(3.8, 0.97515603)
					(4, 0.96711113)
					(4.2, 0.97252487)
					(4.4, 0.969891859999999)
					(4.6, 0.97425543)
					(4.8, 0.95951758)
					(5, 0.97647857)
					(5.2, 0.959936409999999)
					(5.4, 0.95930026)
					(5.6, 0.95775768)
					(5.8, 0.96080141)
					(6, 0.96020308)
					(6.2, 0.958635129999999)
					(6.4, 0.95036463)
					(6.6, 0.956046799999999)
					(6.8, 0.95646436)
					(7, 0.95691725)
					(7.2, 0.95069407)
					(7.4, 0.95124591)
					(7.6, 0.94355364)
					(7.8, 0.954544029999999)
					(8, 0.94985336)
					(8.2, 0.95696104)
					(8.4, 0.95678506)
					(8.6, 0.94879252)
					(8.8, 0.94928524)
					(9, 0.94876937)
					(9.2, 0.94635265)
					(9.4, 0.947944059999999)
					(9.6, 0.94861526)
					(9.8, 0.94695092)
					(10, 0.94016734)
				};
				\addlegendentry{Transformer Sample Size=30}
				\addplot[
				color=black,
				mark=circle,
				]
				coordinates {
					(0.2,0.9315780894477095)
					(10, 0.9315780894477095)
				};
				\addlegendentry{MLE Sample Size=30}
				\addplot[
				color=red,
				mark=circle,
				]
				coordinates {
					(0.2, 0.55541321)
					(0.4, 0.425331777)
					(0.6, 0.387238066)
					(0.8, 0.363779830999999)
					(1, 0.352328115)
					(1.2, 0.338995693)
					(1.4, 0.329424199)
					(1.6, 0.326160580999999)
					(1.8, 0.324902771)
					(2, 0.321227414)
					(2.2, 0.316265022)
					(2.4, 0.313296026999999)
					(2.6, 0.310943053)
					(2.8, 0.310345264999999)
					(3, 0.309173085)
					(3.2, 0.306787439)
					(3.4, 0.307252883)
					(3.6, 0.303599227)
					(3.8, 0.304195398)
					(4, 0.302629938999999)
					(4.2, 0.299885104)
					(4.4, 0.298021443)
					(4.6, 0.300407714)
					(4.8, 0.299224127999999)
					(5, 0.295469004)
					(5.2, 0.295773538)
					(5.4, 0.295311440999999)
					(5.6, 0.296728195)
					(5.8, 0.295223003999999)
					(6, 0.298125782)
					(6.2, 0.294009177)
					(6.4, 0.297182908)
					(6.6, 0.291350135999999)
					(6.8, 0.294836175)
					(7, 0.290738935999999)
					(7.2, 0.288674491)
					(7.4, 0.291823871)
					(7.6, 0.289479128)
					(7.8, 0.289288468)
					(8, 0.287960171999999)
					(8.2, 0.285640788)
					(8.4, 0.28758603)
					(8.6, 0.286891438)
					(8.8, 0.288379394)
					(9, 0.289124325)
					(9.2, 0.285681472)
					(9.4, 0.286081468)
					(9.6, 0.286491802)
					(9.8, 0.28410576)
					(10, 0.283533095)
					
				};
				\addlegendentry{Transformer Sample Size=100}
				\addplot[
				color=red,
				mark=circle,
				]
				coordinates {
					(0.2, 0.27795503440591934)
					(10,  0.27795503440591934)
				};
				\addlegendentry{MLE Sample Size=100}
				\addplot[
				color=brown,
				mark=circle,
				]
				coordinates {
					(0.2, 1.67978351)
					(0.4, 1.38708761999999)
					(0.6, 1.31929418)
					(0.8, 1.30193842)
					(1, 1.27040139)
					(1.2, 1.27071889999999)
					(1.4, 1.23040661)
					(1.6, 1.23863994999999)
					(1.8, 1.21207299)
					(2, 1.22213047)
					(2.2, 1.21174789)
					(2.4, 1.20849301999999)
					(2.6, 1.20101151999999)
					(2.8, 1.20330582999999)
					(3, 1.18102083)
					(3.2, 1.19242699)
					(3.4, 1.19114154)
					(3.6, 1.19342462999999)
					(3.8, 1.17175135)
					(4, 1.17075322)
					(4.2, 1.18380053999999)
					(4.4, 1.18016859999999)
					(4.6, 1.16414584)
					(4.8, 1.1707085)
					(5, 1.16847447999999)
					(5.2, 1.16152839999999)
					(5.4, 1.16650696999999)
					(5.6, 1.16552214)
					(5.8, 1.17508336)
					(6, 1.18298695)
					(6.2, 1.1512782)
					(6.4, 1.1747037)
					(6.6, 1.15118005999999)
					(6.8, 1.16281036)
					(7, 1.15938886)
					(7.2, 1.18090135)
					(7.4, 1.16583680999999)
					(7.6, 1.16188302)
					(7.8, 1.15005252999999)
					(8, 1.14441915)
					(8.2, 1.15367321)
					(8.4, 1.15036657)
					(8.6, 1.144034)
					(8.8, 1.14915073999999)
					(9, 1.15365281)
					(9.2, 1.14429482)
					(9.4, 1.15619894)
					(9.6, 1.14841104)
					(9.8, 1.15480296)
					(10, 1.13756487)
				};
				\addlegendentry{Transformer Sample Size=10-100}
				\addplot[
				color=brown,
				mark=circle,
				]
				coordinates {
					(0.2, 1.1308702929069274)
					(10,  1.1308702929069274)
				};
				\addlegendentry{MLE Sample Size=10-100}
			\end{axis}
		\end{tikzpicture}
	}
	(b)
\end{minipage}
	\caption{The mean-square-error with \#training examples for \textbf{normal} distributions with \textbf{known} parameter ranges. The horizontal lines represent mean-square-errors of MLE, and the curves represent those of our approach.}
	\vspace{-0.2in}
\end{figure}
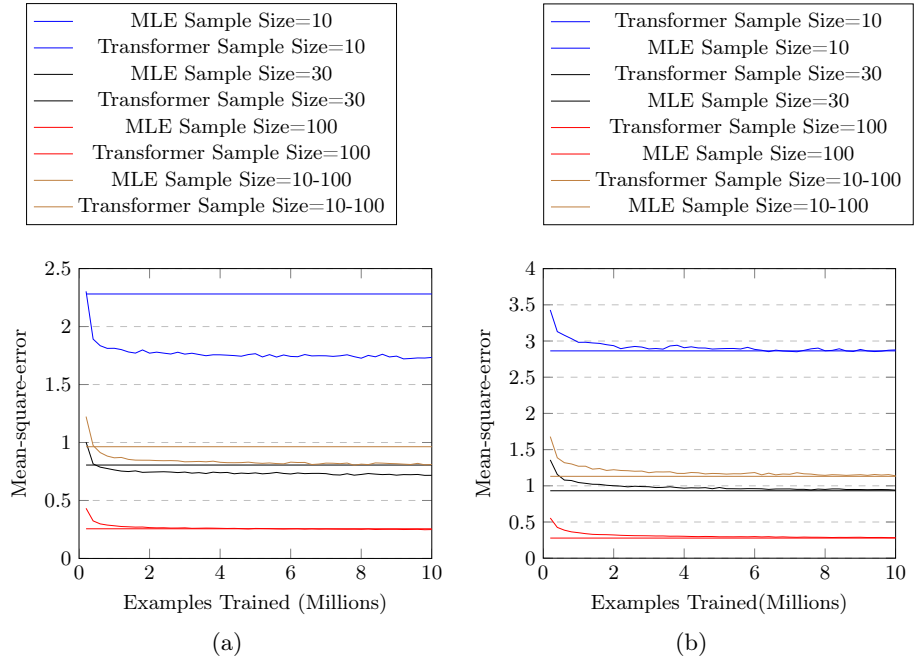

The results are shown in Figure 4(a), and Table 4 shows the mean and standard deviation of the mean-square-error of the two approaches for each sample size. One can see that our approach solidly beats maximum likelihood estimation in every sample size (with $p$-value close to zero in two-sample $t$-tests), especially when the sample size is small. 

\begin{table}[]
	\vspace{-0.1in}
	\begin{tabular}{|p{0.65in}|p{1in}|p{1.5in}|p{0.6in}|p{0.6in}|}\hline
		Sample size & MLE (mean/std) & Our Approach (mean/std) & $t$-value & $p$-value\\\hline
		10 & 2.2809 / 3.3162 & 1.7337 / 2.3722 & 42.439 & 0.0001\\\hline
		30 & 0.8056 / 1.2578 & 0.7169 / 1.0949 & 16.8203 & 0.0001\\\hline
		100 & 0.2558 / 0.3960 & 0.2492 / 0.3786 & 3.8418 & 0.0001\\\hline
		10 to 100 & 0.9636 / 1.7916 & 0.8106 / 1.4343 & 21.0818 & 0.0001\\\hline
	\end{tabular}
	\vspace{0.1in}
	\caption{The mean and standard deviation of the mean-square-error of MLE and our approach for each sample size for \textbf{normal} distribution with \textbf{known} parameter range, together with $t$-value and $p$-value from a two-sample $t$-test. }
	\vspace{-0.2in}
\end{table}

\subsubsection*{4.3.2 Unknown Parameter Range}
Suppose a sample is drawn from a normal distribution $\mathcal{N}(\mu,\sigma^2)$, and the ranges of $\mu$ and $\sigma$ are unknown. Like the case for exponential distribution, we first normalize each sample $s$ into range $[0, 1]$, in order to hide the parameters’ range from our model. Let $a = min(s)$ and $b = max(s)$. Each value $x_i$ is normalized by  $x_i’ = ( x_i - a ) / (b - a)$.

Suppose the model’s outputs are $\mu^{*}$ and $\sigma^{*}$. To compute the loss, we first recover them into the original range: The estimate of $\mu$ is $\hat{\mu} = (\mu^{*} \cdot (b - a)) + a$, and that of $\sigma$ is $\hat{\sigma} = \sigma^{*} \cdot (b - a)$. Then we compare them with the true parameters, as follows:

\begin{equation}
	loss = \mbox{mean-square-error}([\hat{\mu}, \hat{\sigma}], [\mu, \sigma])
\end{equation}

The results are shown in Figure 4(b) and Table 5. One can see that our proposed approach has similar performance (no statistical significance) with MLE in most cases, but is not as good when sample size is 100.

\begin{table}[]
	\vspace{-0.2in}
	\begin{tabular}{|p{0.65in}|p{1in}|p{1.5in}|p{0.6in}|p{0.6in}|}\hline
		Sample size & MLE (mean/std) & Our Approach (mean/std) & $t$-value & $p$-value\\\hline
		10 & 2.8637 / 4.4314&2.8768 / 4.4532&-0.6594&0.5096\\\hline
		30 & 0.9316 / 1.4520&0.9402 / 1.4704&-1.3160&0.1882\\\hline
		100 & 0.2780 / 0.4357&0.0182 / 0.0323&-2.7978&0.0051\\\hline
		10 to 100 & 1.1309 / 2.2248&1.1376 / 2.2967&-0.6626&0.5076\\\hline
	\end{tabular}
	\vspace{0.1in}
	\caption{The mean-square-error of MLE and our approach for each sample size for \textbf{normal} distribution with \textbf{unknown} parameter range, together two-sample $t$-test results. }
	\vspace{-0.5in}
\end{table}

\subsection*{4.4 Beta Distribution}

There is no closed form solution for maximum likelihood estimation for the parameters of a Beta distribution.  Numerical solutions have been proposed (e.g.,~\cite{c:08}), which are complex and there is no open-sourced implementation available. In the special case where the two parameters $\alpha$ and $\beta$ are between 0 and 1, one can use the method of moments~\cite{u:20} to estimate the parameters, and we compare that to our approach. 

\begin{table}[]
	\vspace{-0.1in}
	\begin{tabular}{|p{0.65in}|p{1in}|p{1.5in}|p{0.6in}|p{0.6in}|}\hline
		Sample size & MLE (mean/std) & 	Our Approach (mean/std) &	$t$-value & $p$-value\\\hline
		10 &	0.0933 / 0.0665 &	0.0235 / 0.0261 & 	97.7 &	0.0001\\\hline
		30 &	0.3313 / 0.2086 & 	0.0123 / 0.0155 & 	152.8 &  0.0001\\\hline
		100 & 	0.2873 / 0.1904 &	0.0054 / 0.0081 & 	148.3 &  0.0001\\\hline
		10 to 100 & 	0.1756 / 0.1330 & 	0.0131 / 0.0177 & 	106.4 & 0.0001\\\hline
	\end{tabular}
	\vspace{0.1in}
	\caption{The mean-square-error of MLE and our approach for each sample size for \textbf{Beta} distribution with \textbf{known} parameter range (0 < $\alpha$, $\beta$ < 1), together two-sample $t$-test results. }
\end{table}
\vspace{-0.3in}

Table 6 shows our results of parameter estimation for Beta distribution, compared with Method of Moments. Even in the special case where 0 < $\alpha$, $\beta$ < 1, our approach outperforms Method of Moments by a large margin. On the other hand, our approach can estimate parameters for Beta distributions of any parameters (results shown in Table 7). 

\begin{table}[]
	\vspace{-0.1in}
	\begin{tabular}{|p{0.8in}|p{0.5in}|p{0.5in}|p{0.5in}|p{0.6in}|}\hline
		Sample size & 10 & 	30 & 100 & 10 to 100 \\\hline
		MSE &	0.543 &	0.287  & 0.122 & 0.302\\\hline
	\end{tabular}
	\vspace{0.1in}
	\caption{The mean-square-error of our approach for each sample size for \textbf{Beta} distribution with \textbf{known} parameter range (0.5 < $\alpha$, $\beta$ < 5) }
\end{table}

We can see that our approach is an ideal solution for distributions without readily known formula for parameter estimation. Traditionally it often takes years of research time to derive a closed form or design an algorithm for parameter estimation of a particular type of distribution. In comparison, our approach only requires 60 hours of machine time to train a model for each type of distribution.

\section*{5 Conclusions}
In this paper we propose a new method for parameter estimation, which converts a sample into a sequence of embeddings which can be consumed by a transformer model. The empirical study shows that the proposed approach outperforms maximum likelihood estimation (in terms of mean-square-error) is most scenarios, especially when the parameters’ ranges are known. In real-world applications the parameters’ ranges are usually known, which makes our approach an ideal solution to estimate parameters. 

%
%
%
%

\end{document}